te.tex 
\RequirePackage{fix-cm}
\documentclass[smallextended]{svjour3}       
\smartqed  
\usepackage{graphicx}
\usepackage{ctable}
\usepackage{multirow}
\usepackage{color}
\usepackage{longtable}
\usepackage{subfigure}
\usepackage{fancyhdr}

\fancypagestyle{alim}{
\fancyhf{}

\fancyfoot[CE,CO]{* Accepted at IPCAI 2021 *}}
\begin{document}

\title{Multi-Task Temporal Convolutional Networks for Joint Recognition of Surgical Phases and Steps in Gastric Bypass Procedures
}

\titlerunning{Multi-Task Temporal Convolutional Networks}        

\author{Sanat Ramesh \and
Diego Dall'Alba \and
Cristians Gonzalez \and
Tong Yu \and
Pietro Mascagni \and
Didier Mutter \and
Jacques Marescaux \and
Paolo Fiorini \and
Nicolas Padoy
}


\institute{Sanat Ramesh \and Diego Dall'Alba \and Paolo Fiorini \at
              Altair Robotics Lab, Department of Computer Science, University of Verona, Verona, Italy\\
              \email{\{sanat.ramesh, diego.dallalba, paolo.fiorini\}@univr.it}
           \and
           Sanat Ramesh \and Tong Yu \and Pietro Mascagni \and Nicolas Padoy \at
              ICube, University of Strasbourg, CNRS, IHU Strasbourg, France \\
              \email{\{tyu, p.mascagni, npadoy\}@unistra.fr}
            \and 
            Cristians Gonzalez \at
              University Hospital of Strasbourg, IHU Strasbourg, France
            \and
            Didier Mutter \at
              University Hospital of Strasbourg, IHU Strasbourg, IRCAD, France 
            \and
            Jacques Marescaux \at
              IRCAD, France
            \and
            Pietro Mascagni \at 
              Fondazione Policlinico Universitario Agostino Gemelli IRCCS, Rome, Italy
}


\maketitle
\thispagestyle{alim}

\begin{abstract} \quad

\paragraph{Purpose} 
Automatic segmentation and classification of surgical activity is crucial for providing advanced support in computer-assisted interventions and autonomous functionalities in robot-assisted surgeries. Prior works have focused on recognizing either coarse activities, such as phases, or fine-grained activities, such as gestures. This work aims at jointly recognizing two complementary levels of granularity directly from videos, namely phases and steps.
\paragraph{Methods} We introduce two correlated surgical activities, phases and steps, for laparoscopic gastric bypass procedure. We propose a Multi-task Multi-Stage Temporal Convolutional Network (MTMS-TCN) along with a multi-task Convolutional Neural Network (CNN) training setup to jointly predict the phases and steps and benefit from their complementarity to better evaluate the execution of the procedure. We evaluate the proposed method on a large video dataset consisting of 40 surgical procedures ({\itshape Bypass40}).
\paragraph{Results} We present experimental results from several baseline models for both phase and step recognition on the {\itshape Bypass40} dataset. The proposed MTMS-TCN method outperforms in both phase and step recognition by 1-2\% in accuracy, precision and recall, compared to single-task methods. Furthermore, for step recognition, MTMS-TCN achieves superior performance  of 3-6\% compared to LSTM based models in accuracy, precision and recall.
\paragraph{Conclusion} 
In this work, we present a multi-task multi-stage temporal convolutional network for surgical activity recognition, which shows improved results compared to single-task models on {\itshape Bypass40} gastric bypass dataset with multi-level annotations. The proposed method shows that the joint modeling of phases and steps is beneficial to improve the overall recognition of each type of activity.

\end{abstract}

\section{Introduction}

Recent works in computer-assisted interventions and robot-assisted minimally invasive surgery have seen significant progress in developing advanced support technologies for the demanding scenarios of a modern Operating Room (OR) \cite{Cleary2005495,MaierHein2017,8880624}. Automatic surgical workflow analysis, i.e. reliable recognition of the current surgical activities, plays an important role in the OR by providing the semantic information needed to design assistance systems that can support clinical decision, report generation, and data annotation. This information is at the core of the cognitive understanding of the surgery and could help reduce surgical errors, increase patient safety, and establish efficient and effective communication protocols \cite{BriconSouf2007,Kranzfelder2012,MaierHein2017,8880624}. 

A surgical procedure can be decomposed into activities at different levels of granularity, such as the whole procedure, phases, stages, steps, and actions \cite{Kati2015}. Recent works have strongly focused on developing methods to recognize one specific level of granularity from video data. The visual detection of phases \cite{Twinanda2017,Zisimopoulos2018DeepPhaseSP,Jin2018,Jin2020MultiTaskRC,Czempiel2020TeCNOSP}, robotic gestures \cite{Varadarajan2009,Zappella2013,Ahmidi2017,FunkeBOBWS19} and instruments \cite{AlHajj2018,Jin2018ToolDA,Nwoye2019WeaklySC,Jin2020MultiTaskRC} have for instance seen a surge in research activities, due to their potential impact on developing intra- and post-operative tools for the purposes of monitoring safety, assessing skills, and reporting. Many of these previous works have focused on endoscopic cholecystectomy procedures, utilizing the publicly available large-scale Cholec80 dataset \cite{Twinanda2017}, and on cataract surgical procedures, utilizing the popular CATARACTS dataset \cite{AlHajj2018,Zisimopoulos2018DeepPhaseSP}. 

In this work, we target another type of high volume procedure, namely the gastric bypass. This procedure is particularly interesting for activity analysis as it exhibits a very complex workflow. Gastric bypass is a procedure to treat obesity, which is considered a global health epidemic by the World Health Organization \cite{pmid11234459}, with approximately 500,000 laparoscopic bariatric procedures performed every year worldwide \cite{Angrisani2015}. Laparoscopic Roux-En-Y gastric bypass (LRYGB), the most performed and gold standard bariatric surgical procedure \cite{Angrisani2015}, consists in the reduction of the stomach and the bypass of part of the small bowel. Various clinical groups have worked to find a consensus on the best workflow for this technically demanding surgical procedure in order to improve standardization and reproducibility \cite{Kaijser2018}. A clear framework and shared nomenclature to segment surgical procedures is currently missing. 

Similar to \cite{Kaijser2018}, we introduce a hierarchical representation of LRYGB procedure containing phases and steps representing the workflow performed in our hospital, and focus our attention on the recognition of these two types of activities. Towards this end, we introduce a new large-scale dataset, called {\itshape Bypass40}, containing 40 endoscopic videos of gastric bypass surgical procedures annotated with  both phases and steps. Overall, 11 phases and 44 steps are annotated in all videos. To the best of our knowledge, this is the first dataset with hierarchical annotations presented for recognition directly from video data. This opens new possibilities for research in surgical knowledge modeling and recognition. To jointly learn the tasks of phase and step recognition, we introduce MTMS-TCN, a Multi-Task Multi-Stage Temporal Convolutional Networks, extending MS-TCNs \cite{farha2019mstcn} proposed for action segmentation.

The contributions of this paper are threefold: (1) we introduce new multi-level surgical activity annotations for the LRYGB procedure and utilize a novel dataset; (2) we propose a multi-task recognition model utilizing only visual features from the endoscopic video; and (3) we benchmark the proposed method with other state-of-the-art deep learning models on the new {\itshape Bypass40} dataset for surgical activity recognition, demonstrating the effectiveness of the joint modeling of phases and steps.

\section{Related work}
\begin{figure}
\centering
\includegraphics[width=\textwidth]{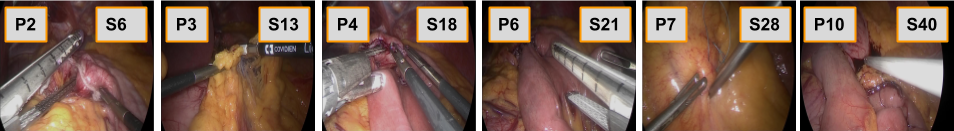}
\caption{Sample images from the dataset with phase labels on top-left and step labels on top-right corner. The labels can be inferred from Tables \ref{tab1:phases} and \ref{tab2:steps}.}. \label{fig1:imgs}
\end{figure}
\setlength{\tabcolsep}{6pt}
\begin{table}
\centering
\caption{List of the phases defined in the dataset with the surgically critical phases highlighted in bold.}\label{tab1:phases}
\begin{tabular}{ r  c }
\hline\noalign{\smallskip}
Phase ID & Phase\\
\noalign{\smallskip}\hline\noalign{\smallskip}
P1 &  preparation\\
{\bfseries P2} &  {\bfseries gastric pouch creation}\\
P3 &  omentum division\\
{\bfseries P4} &  {\bfseries gastrojejunal anastomosis}\\
{\bfseries P5} &  {\bfseries anastomosis test}\\
P6 &  jejunal separation\\
P7 &  closure petersen space\\
{\bfseries P8} &  {\bfseries jejunojejunal anastomosis}\\
P9 &  closure mesenteric defect\\
P10 &  cleaning coagulation\\
P11 &  disassembling\\
\noalign{\smallskip}\hline
\end{tabular}
\end{table}
EndoNet \cite{Twinanda2017} and DeepPhase \cite{Zisimopoulos2018DeepPhaseSP} belong to the early works that employed deep learning for surgical workflow analysis on cholecystectomy and cataract surgeries. EndoNet jointly performed phase and tool detection, and the model consisted of a CNN followed by a hierarchical Hidden Markov Model (HMM) for modeling temporal information, while DeepPhase used a CNN followed by Recurrent Neural Network (RNN) temporal modeling. EndoNet was evolved to EndoLSTM \cite{Twinanda2017VisionbasedAF} that consisted of a CNN for feature extraction and an LSTM, i.e., Long Short-Term Memory, for temporal refinement. Similarly, SV-RCNet \cite{Jin2018} trained an end-to-end ResNet \cite{He2016} and LSTM model incorporating a prior knowledge inference scheme for surgical phase recognition. MTRCNet-CL \cite{Jin2020MultiTaskRC} proposed a multi-task model to detect tool presence and phase recognition. The features from the CNN were used to detect tool presence, and also served as input to a LSTM model for phase prediction. Additionally, a correlation loss was introduced to enhance the synergy between the two tasks. Most of the previous methods use LSTMs, which retains memory for a limited sequence. Since the average duration of a surgery can range from less than half an hour to many hours, it makes it challenging for LSTM based models to leverage the temporal information for surgical phase recognition. 

Temporal Convolutional Networks (TCNs) \cite{Lea2016} were introduced to hierarchically process videos for action segmentation. An encoder-decoder architecture was able to encode both high- and low-level features in contrast to RNNs. Furthermore, dilated convolutions \cite{45774} were utilized in TCNs for action segmentation that showed performance improvements due to a large receptive field for higher temporal resolution. MS-TCN \cite{farha2019mstcn} consisted of a multi-stage predictor architecture with each stage consisting of multi-layer TCN, which incrementally refined the prediction of the previous stage. Recently, TeCNO \cite{Czempiel2020TeCNOSP} adapted the MS-TCN architecture for online surgical phase prediction by implementing  causal convolutions \cite{45774}. We build upon this architecture and confirm experimentally that it is superior to LSTM for multi-level activity recognition.
\setlength{\tabcolsep}{2pt}
\begin{table}
\centering
\caption{List of the steps defined in the dataset with the surgically critical steps highlighted in bold.}\label{tab2:steps}
\begin{tabular}{r c | r c }
\hline\noalign{\smallskip}
Step ID & Step & Step ID & Step\\
\noalign{\smallskip}\hline\noalign{\smallskip}
S0 &  null step & S22 &  gastric tube placement \\
S1 &  cavity exploration & S23 &  clamping \\
S2 &  trocar placement & S24 &  ink injection \\
S3 &  retractor placement & {\bfseries S25} &  {\bfseries visual assessment} \\
{\bfseries S4} &  {\bfseries crura dissection} & S26 &  gastrojejunal anastomosis reinforcement \\
{\bfseries S5} &  {\bfseries his angle dissection} & S27 &  petersen space exposure \\
{\bfseries S6} &  {\bfseries horizontal stapling} & S28 &  petersen space closing \\
{\bfseries S7} &  {\bfseries retrogastric dissection} & S29 &  biliary limb opening \\
{\bfseries S8} &  {\bfseries vertical stapling} & {\bfseries S30} &  {\bfseries alimentary limb measurement} \\
S9 &  gastric remnant reinforcement & S31 &  alimentary limb opening \\
S10 &  gastric pouch reinforcement & {\bfseries S32} &  {\bfseries jejunojejunal stapling} \\
S11 &  gastric opening & S33 &  jejunojejunal defect closing \\
S12 &  omental lifting & S34 &  jejunojejunal anastomosis reinforcement \\
S13 &  omental section & S35 &  staple line reinforcement \\
S14 &  adhesiolysis & S36 &  mesenteric defect exposure \\
S15 &  treitz angle identification & S37 &  mesenteric defect closing \\
{\bfseries S16} &  {\bfseries biliary limb measurement} & S38 &  anastomosis fixation \\
S17 &  jejunum opening & {\bfseries S39} &  {\bfseries coagulation} \\
{\bfseries S18} &  {\bfseries gastrojejunal stapling} & S40 &  irrigation aspiration \\
S19 &  gastrojejunal defect closing & S41 &  parietal closure \\
S20 &  mesenteric opening & S42 &  trocar removal \\
S21 &  jejunal section & S43 & calibration \\
\noalign{\smallskip}\hline
\end{tabular}
\end{table}
\section{Hierarchical Surgical Activities: Phases \& Steps}\label{section:dataset}

We introduce two hierarchically defined surgical activities called phases and steps for the LRYGB procedure. These two elements define the workflow of the surgery at two levels of granularity with the phases describing the surgical workflow at coarser level than the steps. Phases describe a set of fundamental surgical aims to accomplish in order to successfully complete the surgical procedure, while steps describe a set of surgical actions to perform in order to accomplish a surgical phase. The surgical procedure is segmented into 44 fine-grained steps, along with 11 coarser phases. All the phases and steps are listed in Table \ref{tab1:phases} and Table \ref{tab2:steps}. These two types of activities are interesting for their inherent hierarchical relationship, which is shown in Fig. \ref{fig2:phase_step_relation}. 

\begin{figure}
\centering
\subfigure[]{\includegraphics[width=0.7\textwidth]{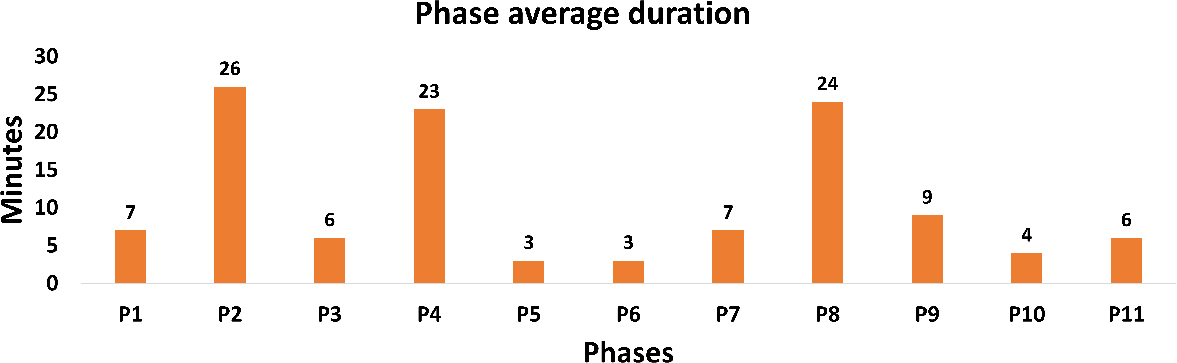}}
\subfigure[]{\includegraphics[width=0.7\textwidth]{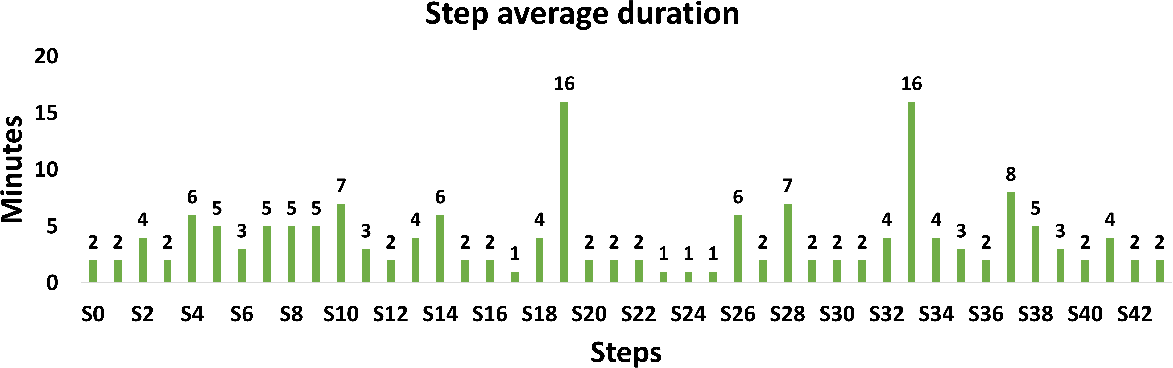}}
\caption{Average duration of phases and steps across videos in the dataset.} \label{fig2:phase_step_dist}
\end{figure}

\begin{figure}
\centering
\includegraphics[width=0.55\textwidth]{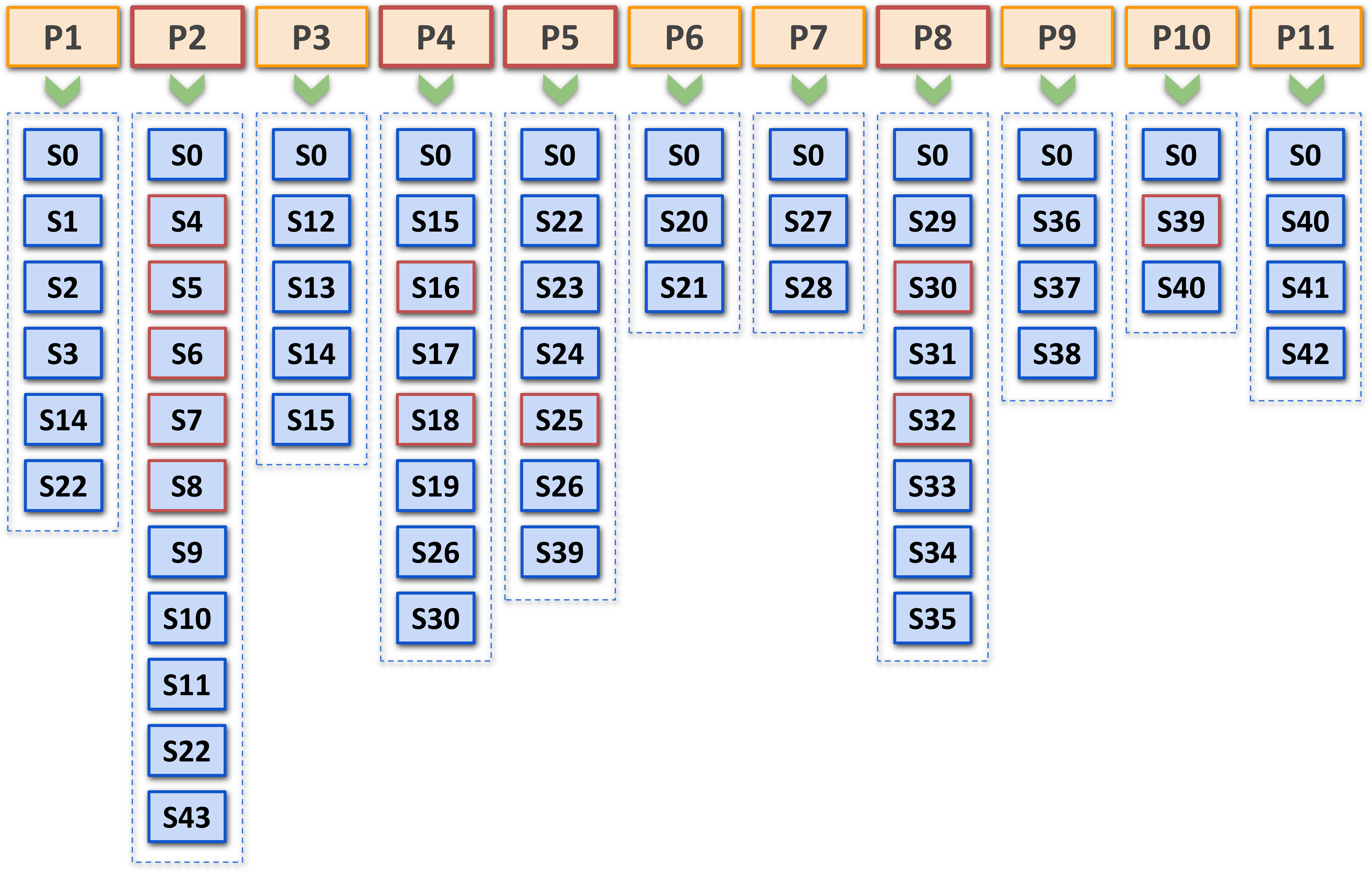}
\caption{Hierarchical relationship between phases and steps with surgically critical activities highlighted in red.} \label{fig2:phase_step_relation}
\end{figure}

\begin{figure}
\centering
{\includegraphics[width=0.9\textwidth]{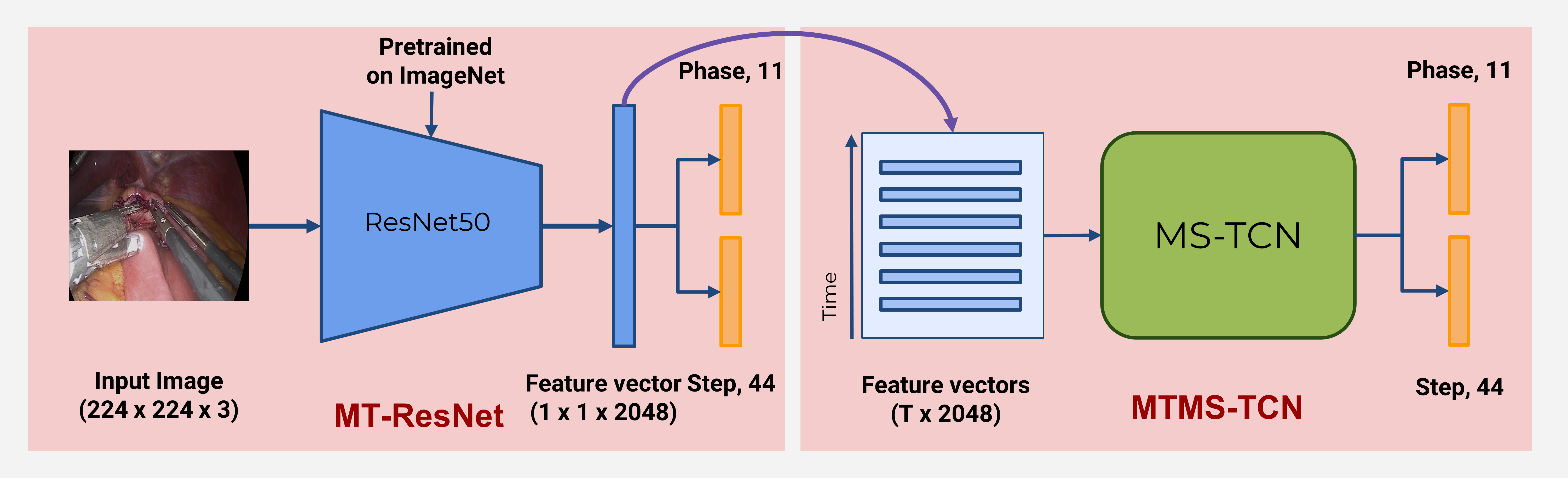}}
\caption{Overview of our model setup. Multi-task architecture of the ResNet-50 feature extractor backbone on the left and the multi-task setup of the TCN temporal model on the right.} \label{fig2:model_arch}
\end{figure}

We introduce a new dataset, called {\itshape Bypass40}, consisting of 40 videos of LRYGB procedures with an average duration of 110$\pm$30 minutes. This dataset is created from surgeries performed by 7 expert surgeons at IHU Strasbourg. The videos are captured at 25 frames-per-second (fps) with a resolution of $854 \times 480$ or $1920 \times 1080$ and annotated with phases and steps.
Sample images with respective phase labels are shown in Fig. \ref{fig1:imgs}. The distribution of phases and steps in the bypass40 dataset is shown in Fig. \ref{fig2:phase_step_dist}. As can be seen, there is a high imbalance in class distribution of both phases and steps. This is to be expected as all steps need not occur in all surgeries and also task completion time of the phases/steps may not be similar. 
%
%
\paragraph{Methodology}
With the aim of joint online recognition of phases and steps, we propose an online surgical activity recognition pipeline consisting of the following steps: 1) A multi-task ResNet-50 is employed as a visual feature extractor. 2) A multi-task multi-stage causal TCN model refines the extracted feature of the current frame by encoding temporal information deduced by analyzing the history. We propose this two-step approach so that the temporal model training is independent of the backbone CNN feature extraction models. The overview of the model setup is depicted in Fig. \ref{fig2:model_arch}.
\paragraph{Feature Extraction Architecture}\label{section:cnn}
ResNet-50 \cite{He2016_V2} has been successfully employed in many works for phase segmentation \cite{DBLP:journals/corr/abs-1812-00033,Czempiel2020TeCNOSP,Jin2018,Jin2020MultiTaskRC}. In this work, we utilize the same architecture as our backbone visual feature extraction model. The model maps $224 \times 224 \times 3$ RGB images to a feature space of size $N_f = 2048$. The model is trained on frames extracted from the videos, without any temporal context, in a multi-task setup to predict both phase and step as shown in Fig. 1 (a). Since both activities are multi-class classification problems that exhibit imbalance in the class distribution, softmax activations and class-weighted cross-entropy loss is utilized. The class weights for both activities are calculated using the median frequency balancing \cite{7410661}.
The total loss, $\mathcal{L}_{total} = \mathcal{L}_{phase} + \mathcal{L}_{step}$\label{eqn1:total_loss}, is obtained by combining equally weighted contributions of class-weighted cross-entropy loss for phases $(\mathcal{L}_{phase})$ and steps $(\mathcal{L}_{step})$.
%
%
\paragraph{Temporal Modeling}
For joint temporal surgical activity recognition task, we propose MTMS-TCN, a multi-task extension of a multi-stage temporal convolutional network. The model takes an input video consisting of $x_{1:t}$, $t \in [1, T]$ frames, where $T$ is the total number of frames, and predicts $y_{1:t}$ where $y_t$ is the class label for the current timestamp $t$. Following the design of MS-TCN, MTMS-TCN contains neither pooling layers nor fully connected layers and it is only constructed with temporal convolutional layers.
Our temporal model consists of only temporal convolutional layers, in particular they are dilated residual layers performing dilated convolutions. Since our aim is to segment surgical activities online, similar to TeCNO \cite{Czempiel2020TeCNOSP}, we perform causal convolutions \cite{45774} at each layer which depends only on $n$ past frames and does not rely on any future frames. The dilation factor is increased by a factor of $2$ for each consecutive layer which increases exponentially the temporal receptive field of the network without introducing any pooling layer. Additionally, the multi-stage model recursively refines the output of the previous stage.

Similar to our setup for CNN, we train our MTMS-TCN in a multi-task fashion to jointly predict the two activities by attaching two heads at the end of a stage. Softmax activations with cross-entropy loss for phase and step is applied and the total loss is similar to the loss utilized for training the backbone CNN (Eq. \ref{eqn1:total_loss}). Please note that the cross-entropy loss is not class-weighted. This is done to allow the temporal model to learn implicitly the duration and occurrence of each class in both phases and steps. 
%
\section{Experimental Setup}
%
\paragraph{Dataset} We evaluate our method on the {\itshape Bypass40} dataset described in Section \ref{section:dataset}. 
We split the 40 videos in the dataset into 4 subsets of 10 videos each to perform 4-fold cross-validation. Each subset was used as test set, while the other subsets were combined together and divided into training and validation tests consisting of 24 and 6 videos respectively. The dataset was subsampled at 1 fps amounting to approximately 149,000 frames for training, 41,000 frames for validation, and 66,000 frames for testing in each fold. The frames are resized to ResNet-50's input dimension of $224 \times 224 \times 3$ and the training dataset is augmented by applying horizontal flip, saturation, and rotation. 
\begin{figure}
\centering
\includegraphics[width=0.95\textwidth]{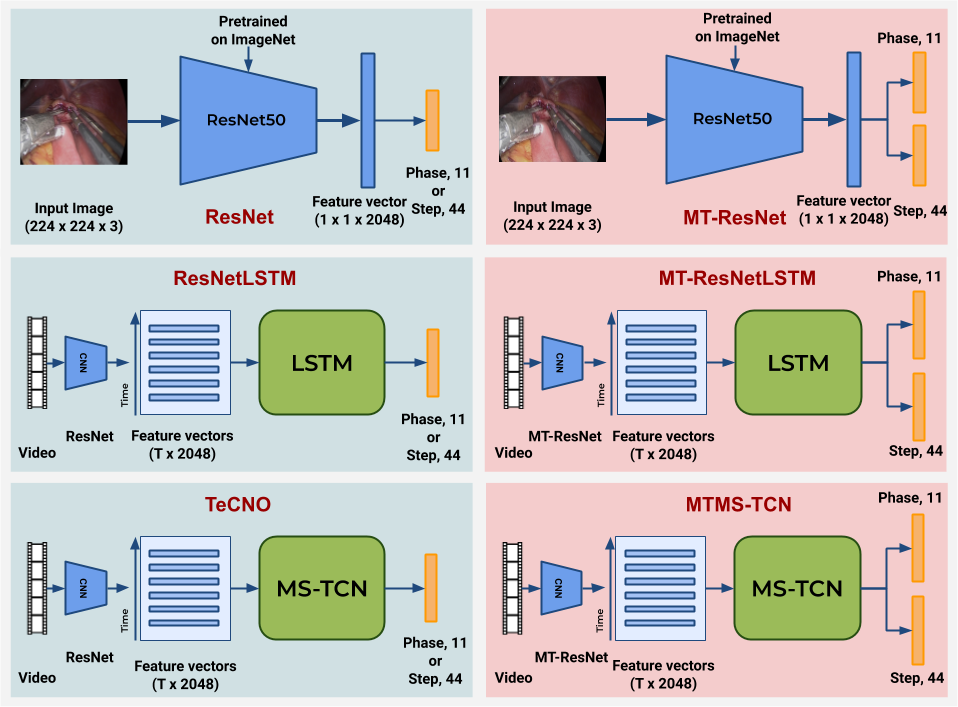}
\caption{Overview of all the models used for evaluation. All the models trained in a single-task setup are shown on the left, while all the models trained in multi-task setup are shown on the right.}\label{fig2:all_model_arch}
\end{figure}
%
%

\paragraph{Model Training} The ResNet-50 model is initialized with weights pre-trained on ImageNet. The model is then trained for the task of phase and step recognition in a single-task setup, called ResNet, and jointly in a multi-task setup, called MT-ResNet, described in Section \ref{section:cnn}. In all the experiments, the model is trained for $30$ epochs with a learning rate of 1e-5, weight regularization of 5e-5, and a batch size of 32. The test results presented are from the best performing model on the validation set. 
The baseline TCN model is trained in a single-task setup utilizing the features extracted from backbone ResNet (Fig. \ref{fig2:all_model_arch}). This is effectively achieved by training TeCNO separately for the two activity recognition tasks. The MTMS-TCN model is trained in a multi-task setup utilizing the backbone MT-ResNet trained in a similar fashion. In all the experiments, the model is trained for 200 epochs with a learning rate of 3e-4. The features representations of augmented data for CNN is also utilized for training the TCN model (Fig. \ref{fig2:all_model_arch}). Our CNN backbone was implemented in Tensorflow while the temporal models (TCN and LSTM) were implemented in Pytorch. Our models were trained on NVIDIA GeForce RTX 2080 Ti GPUs.
%

\paragraph{Evaluation Metrics} We follow the same evaluation metrics used in other related publications  \cite{Czempiel2020TeCNOSP,Jin2018,Jin2020MultiTaskRC}, where accuracy (ACC), precision (PR), recall (RE), and F1 score (F1) are used to effectively compare the results. Accuracy quantifies the total correct classification of activity in the whole video. PR, RE, and F1 are computed class-wise, defined as:
\begin{equation}
    PR=\frac{| GT\cap P |}{| P |},\  RE=\frac{| GT\cap P |}{| GT |},\  F1= \frac{2}{\frac{1}{PR} + \frac{1}{RE}},
\end{equation}
where GT and P represent the ground truth and prediction for one class, respectively. These values are averaged across all the classes to obtain PR, RE, and F1 for the entire test set. We perform 4-fold cross-validation and report the results as mean and standard deviation across all the folds.

\setlength{\tabcolsep}{2.5pt}
\begin{table}
\centering
\caption{Baseline comparison on the dataset for phase recognition. Accuracy (ACC), precision (PR), recall (RE), and F1-score (F1) (\%) are reported  across all the 4-fold cross-validation. }\label{tab1:phase_metrics}
\begin{tabular}{ c c c c c c }

\hline\noalign{\smallskip}
       &      & \multicolumn{4}{c}{Phase}\\
\noalign{\smallskip }\cline{3-6} \noalign{\smallskip} 
       &      &        ACC &          PR &       RE &        F1   \\
\noalign{\smallskip}\hline\noalign{\smallskip}
\multirow{4}{*}{No TCN} & ResNet          &      82.07 $\pm$ 3.31	& 73.89 $\pm$ 3.27	& 72.22 $\pm$ 3.39	& 72.46 $\pm$ 3.60   \\
& MT-ResNet &      81.67 $\pm$ 2.67	 &	73.06 $\pm$ 2.77 &	72.05 $\pm$ 2.25 &	72.13 $\pm$ 2.59   \\
& ResNetLSTM        &      {\bfseries 89.11 $\pm$ 2.75} & 	{\bfseries 82.08 $\pm$ 3.58} & 	{\bfseries 82.25 $\pm$ 3.45} & 	{\bfseries 81.66 $\pm$ 3.52}   \\
& MT-ResNetLSTM      &      88.62 $\pm$ 2.65 &	81.35 $\pm$ 3.94 &	81.09 $\pm$ 3.46 &	80.65 $\pm$ 3.83   \\
\noalign{\smallskip}\hline\noalign{\smallskip}
\multirow{2}{*}{Stage I} & TeCNO        &       89.84 $\pm$3.50 &    85.40 $\pm$	4.02 &    82.29 $\pm$	4.51 &    82.98 $\pm$ 	4.06     \\
& MTMS-TCN       &      {\bfseries 91.16 $\pm$2.92} &	{\bfseries 86.11 $\pm$	3.69} &	{\bfseries 83.80 $\pm$ 4.04} &	{\bfseries 84.44 $\pm$ 3.53}  \\
\noalign{\smallskip}\hline\noalign{\smallskip}
\multirow{2}{*}{Stage II} & TeCNO        &      89.92 $\pm$ 3.25 &	84.39 $\pm$ 4.28	 &	83.34 $\pm$ 3.91 &	83.54 $\pm$ 3.96   \\
& MTMS-TCN       & {\bfseries 90.93 $\pm$ 3.21}  &	{\bfseries 85.60 $\pm$ 4.47}  &	{\bfseries 83.99 $\pm$ 4.19}  &	{\bfseries 84.21 $\pm$ 4.17}   \\
\noalign{\smallskip}\hline
\end{tabular}
\end{table}
\setlength{\tabcolsep}{2.5pt}
\begin{table}
\centering
\caption{Baseline comparison on the dataset for step recognition. Accuracy (ACC), precision (PR), recall (RE), and F1-score (F1) (\%) are reported  across all the 4-fold cross-validation. }\label{tab1:step_metrics}
\begin{tabular}{ c c c c c c c c c c c }

\hline\noalign{\smallskip}
       &      & \multicolumn{4}{c}{Step} \\
\noalign{\smallskip }\cline{3-6} \noalign{\smallskip} 
       &      &        ACC &          PR &       RE &        F1   \\
\noalign{\smallskip}\hline\noalign{\smallskip}
\multirow{4}{*}{No TCN} & ResNet          &       65.53 $\pm$ 2.04 &	45.31 $\pm$ 3.02 &	43.24 $\pm$ 2.71 &	42.62 $\pm$ 2.26 \\
& MT-ResNet &      66.63 $\pm$ 2.37	 &	45.96 $\pm$ 3.12 &	44.67 $\pm$ 3.13 &	43.80 $\pm$ 2.86 \\
& ResNetLSTM        &      71.34 $\pm$ 2.25 &	47.79 $\pm$ 4.13 &	47.67 $\pm$ 2.80 &	45.82 $\pm$ 2.66   \\
& MT-ResNetLSTM      &      {\bfseries 72.21 $\pm$ 2.02}  &	{\bfseries 50.97 $\pm$ 3.32}  &	{\bfseries 49.25 $\pm$ 1.80}  &	{\bfseries 47.87 $\pm$ 2.09} \\
\noalign{\smallskip}\hline\noalign{\smallskip}
\multirow{2}{*}{Stage I} & TeCNO        &      75.12 $\pm$ 2.43 &   54.69 $\pm$ 2.56	&   50.91 $\pm$ 2.42		&   49.91 $\pm$ 1.76     \\
& MTMS-TCN       &      {\bfseries 76.09 $\pm$2.67}    &	{\bfseries 56.44 $\pm$	3.60}	&	{\bfseries 52.51 $\pm$	3.28} &	{\bfseries 51.89 $\pm$	2.90}  \\
\noalign{\smallskip}\hline\noalign{\smallskip}
\multirow{2}{*}{Stage II} & TeCNO        &      74.78 $\pm$ 2.48 &	53.15 $\pm$ 2.47  &	50.76 $\pm$ 3.27  &	49.86 $\pm$ 3.65  \\
& MTMS-TCN       &      {\bfseries 75.45 $\pm$ 3.11}  &	{\bfseries 54.88 $\pm$ 	4.35}  &	{\bfseries 52.59 $\pm$ 4.20}  &	{\bfseries 51.77 $\pm$ 4.10}  \\
\noalign{\smallskip}\hline
\end{tabular}
\end{table}
%
%

The overview of all evaluated models is depicted in Fig. \ref{fig2:all_model_arch}. MTMS-TCN is evaluated against popular surgical phase recognition networks, ResNetLSTM \cite{Jin2018}, and TeCNO \cite{Czempiel2020TeCNOSP}. Both these networks are trained in a two-step process for the single-task of phase and step separately. Furthermore, ResNetLSTM is extended to get MT-ResNetLSTM where the ResNetLSTM model is trained in a multi-task setup. Since causal convolutions are used in the model for online recognition of activities, for fair comparison unidirectional LSTM is utilized. The LSTM, with 64 hidden units, is trained using the video features extracted from the CNN backbone with a sequence length equal to the length of the videos for $200$ epochs with a learning rate of 3e-4.
\section{Results and Discussions}
%
%
%
\begin{figure}
\centering
\includegraphics[width=9cm]{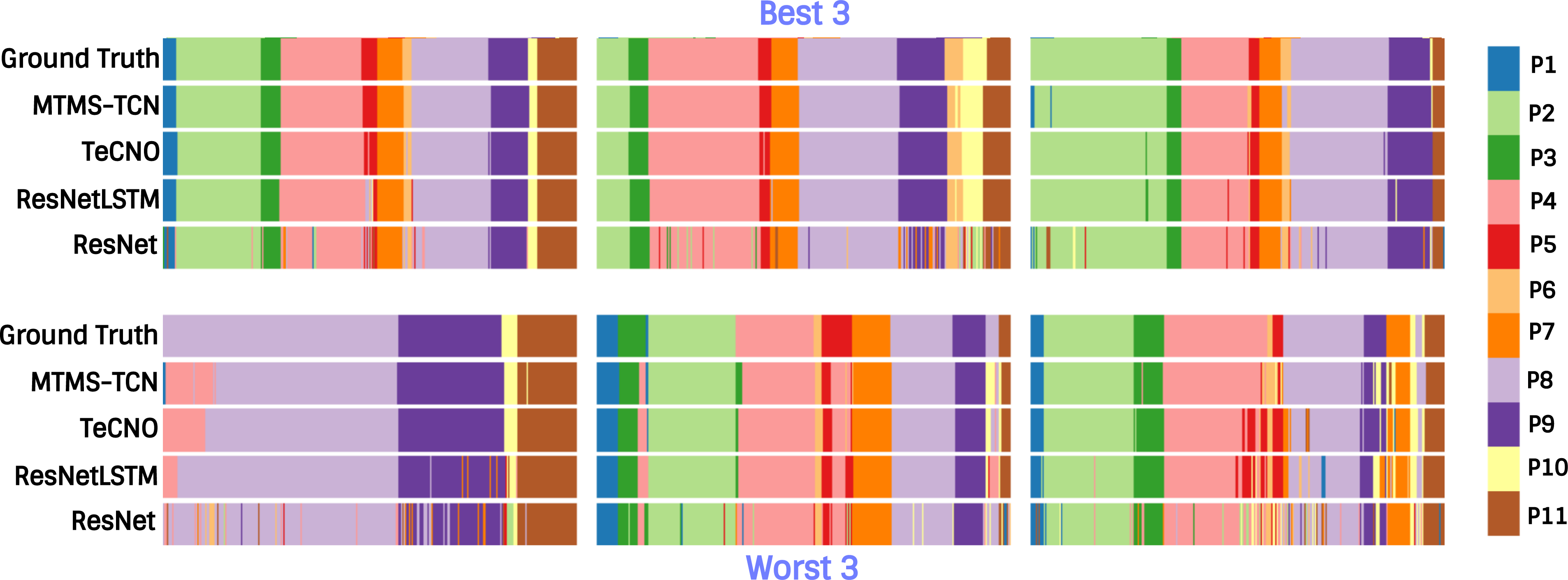}
\caption{Phase recognition on complete videos in Bypass40 for quality assessment. The top row shows 3 videos on which our model performs best and bottom row shows 3 videos with worst performance.} \label{fig:phase_predictions}
\end{figure}
\begin{figure}
\centering
\includegraphics[width=9.5cm]{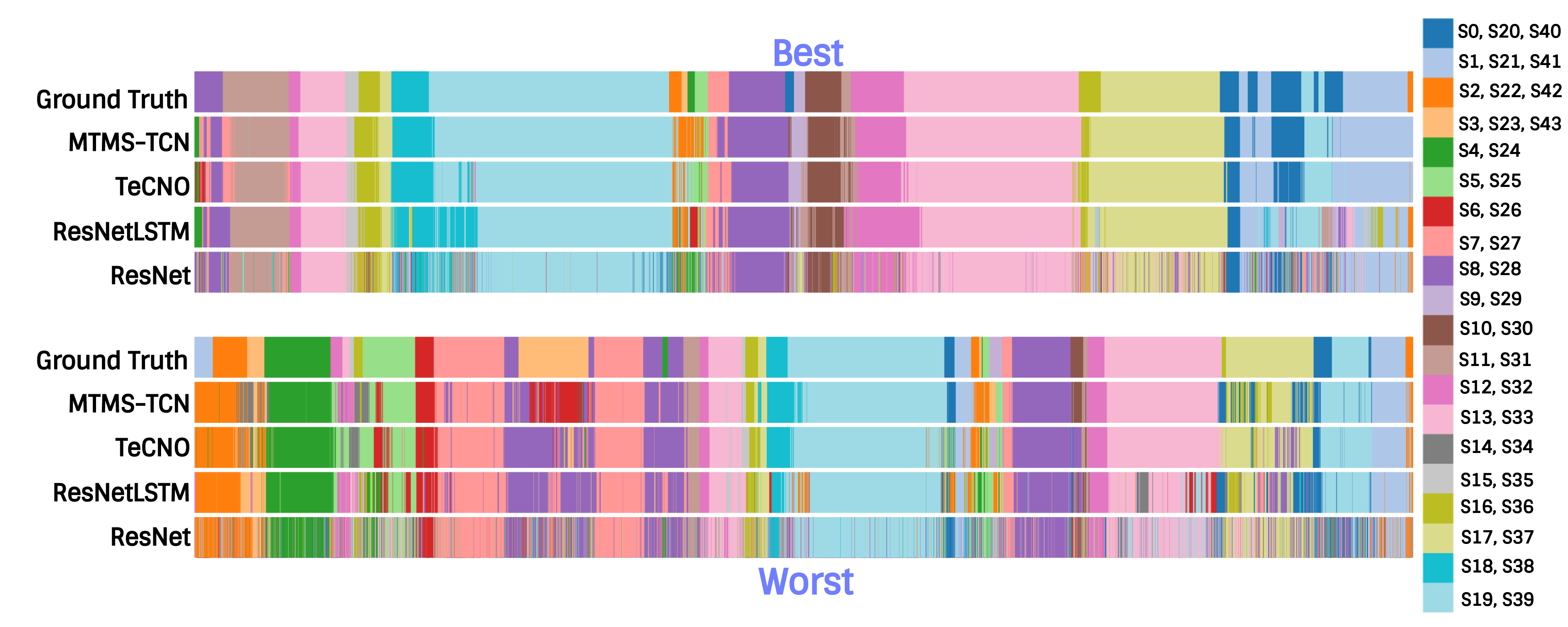}
\caption{Step recognition on complete videos in Bypass40 for quality assessment. The figure shows best (top) and worst (bottom) performance of our model. The 44 distinct steps are mapped to the same 20 categorical colormap.} \label{fig:step_predictions}
\end{figure}

\setlength{\tabcolsep}{2.5pt}
\begin{table}
\centering
\caption{Baseline comparison on the dataset for joint phase and step recognition. Accuracy (ACC) is reported  after 4-fold cross-validation}\label{tab1:joint_metrics}
\begin{tabular}{ c c c c c }
\noalign{\smallskip}\hline\noalign{\smallskip}
      &      &      Phase ACC &     Step ACC &  Phase-Step ACC \\
\noalign{\smallskip}\hline\noalign{\smallskip}
\multirow{2}{*}{No TCN} 
& ResNet & 82.07 $\pm$ 2.86 & 65.53 $\pm$ 1.76 & 54.87 $\pm$ 2.55 \\
& MT-ResNet &     81.67 $\pm$ 2.31 & 66.63 $\pm$ 2.05 & 64.77 $\pm$ 2.01 \\
& ResNetLSTM & {\bfseries 89.11 $\pm$ 2.38} & 71.33 $\pm$ 1.95 & 68.48 $\pm$ 2.33 \\
& MT-ResNetLSTM      &      88.62 $\pm$ 2.29 & {\bfseries 72.21 $\pm$ 1.75} & {\bfseries 70.69 $\pm$ 1.92} \\
\noalign{\smallskip}\hline\noalign{\smallskip}
Stage I 
& TeCNO          &   89.84 $\pm$ 3.03 & 75.12 $\pm$ 2.10 & 72.32 $\pm$ 2.95 \\
& MTMS-TCN       &     {\bfseries 91.16 $\pm$ 2.53} & {\bfseries 76.09 $\pm$ 2.32} & {\bfseries 75.14 $\pm$ 2.76} \\
\noalign{\smallskip}\hline\noalign{\smallskip}
Stage II 
& TeCNO & 89.91 $\pm$ 2.81 & 74.78 $\pm$ 2.15 & 71.89 $\pm$ 2.71 \\
& MTMS-TCN       & {\bfseries 90.93 $\pm$ 2.78} & {\bfseries 75.45 $\pm$ 2.69} & {\bfseries 75.06 $\pm$ 2.81} \\
\noalign{\smallskip}\hline
\end{tabular}
\end{table}

Comparison of MTMS-TCN (Stage I) with other state-of-the-art methods, utilizing both LSTMs and TCNs, is presented in Table \ref{tab1:phase_metrics} and Table \ref{tab1:step_metrics} on both phase and step recognition tasks. TeCNO which utilizes TCNs outperforms both ResNetLSTM and MT-ResNetLSTM models by 1\% and 3\% in terms of accuracy. MTMS-TCN outperforms TeCNO, ResNetLSTM and MT-ResNetLSTM models for by 2\% the phase recognition. 

Similarly, for step recognition, TeCNO outperforms both LSTM based models by 3-4\% with respect to Accuracy, and 3-6\% in terms of precision. MTMS-TCN improves over TeCNO by 1\% in accuracy and outperforms it by 2\% and 1.5\% in terms of precision and recall, respectively. In turn, MTMS-TCN outperforms LSTM based models by 4-5\% in terms of accuracy and 3-8\% in terms of precision and recall. 

Table \ref{tab1:joint_metrics} presents performance of all the models on joint recognition of phase and step. We present joint phase-step prediction accuracy which is computed as the average number of instances where both the phase and step are correctly recognized by the model. All the multi-task models outperform their single-task counterpart. In particular, MTMS-TCN outperforms TeCNO by 3\%. Moreover, the joint-recognition accuracy of MTMS-TCN is very close to its step recognition accuracy which indicated that the model has implicitly learned the hierarchical relationship and benefited from it. 

The improvement achieved by both MTMS-TCN and TeCNO in both the recognition tasks over LSTM based models is attributed to the higher temporal resolution and large receptive field of the underlying TCN module. On the other hand, improvement of MTMS-TCN over TeCNO is attributed to the multi-task setup. Additionally, MT-ResNet, the backbone of our MTMS-TCN, achieves improved performance in steps with a small decrease in performance for phase recognition compared to ResNet, the backbone of TeCNO. 

A set of surgically critical steps along with their average precision, recall, and F1-score is presented in Table \ref{tab1:per_step_metrics}. MTMS-TCN performs  better than TeCNO in recognizing many of the steps. Moreover, short duration steps such as S25, S30, and S39 that are harder to recognize, are significantly better recognized by our MTMS-TCN over TeCNO. All these results validates our model trained in a multi-task setup for joint recognition of phases and steps.

\setlength{\tabcolsep}{6pt}
\begin{table}
\centering
\caption{TeCNO vs MTMS-TCN: 4-fold cross-validation  average precision, recall, and F1-score (\%) reported for the critical steps.}\label{tab1:per_step_metrics}
\begin{tabular}{r c c c c c c c}
\hline\noalign{\smallskip}
    & \multicolumn{3}{c}{TeCNO} && \multicolumn{3}{c}{MTMS-TCN} \\
\noalign{\smallskip }\cline{2-4} \cline{6-8} \noalign{\smallskip}
Steps ID    &   PR &   RE &   F1  &&   PR &   RE &   F1 \\
\noalign{\smallskip}\hline\noalign{\smallskip}
  S4 &       84.47 &    {\bfseries 90.07} &    85.77 &&       {\bfseries 86.76} &    88.44 &    {\bfseries 86.39} \\
  S5 &       {\bfseries 87.54}  &    {\bfseries 80.32}  &    80.59 &&       87.07 &   77.16  &    78.66 \\
  S6 &       79.02  &   62.04  &    62.05 &&       {\bfseries 79.45}  &    {\bfseries 63.47} &    {\bfseries 62.41} \\
  S7 &       {\bfseries 78.76} &    {\bfseries 65.87} &    {\bfseries 69.23} &&       73.43  &     65.66 &     67.92 \\
  S8 &       {\bfseries 78.10}  &    {\bfseries 77.02} &    {\bfseries 72.76} &&       75.59 &   {\bfseries 77.18}  &    {\bfseries 72.73} \\
 S16 &       76.48 &    {\bfseries 68.83} &    {\bfseries 68.60} &&       {\bfseries 79.09} &    67.47 &    {\bfseries 68.41} \\
 S18 &       {\bfseries 92.47} &    {\bfseries 82.81} &    {\bfseries 86.44} &&       89.95 &     80.37 &     83.44 \\
 S25 &       {\bfseries 55.51} &    39.73 &    40.88 &&       47.28 &    {\bfseries 49.50} &    {\bfseries 44.98} \\
 S30 &       62.03 &    63.48 &    58.22 &&  {\bfseries 64.02}      &    {\bfseries 70.90} &    {\bfseries 63.76}  \\
 S32 &       {\bfseries 88.14} &    85.33 &    {\bfseries 83.92} &&       85.72 &    {\bfseries 86.13} &    {\bfseries 83.87} \\
 S39 &       55.36 &   41.78  & 38.86 &&       {\bfseries 62.69} &    {\bfseries 48.12} &    {\bfseries 48.85} \\
\noalign{\smallskip}\hline
\end{tabular}
\end{table}
%
%
Fig. \ref{fig:phase_predictions} visualizes a video set of 3 best and 3 worst performance of MTMS-TCN for phase recognition. The predictions of MTMS-TCN in some cases performs better than TeCNO in recognizing smaller phases, such as P5, P7, P9 and P10.
MTMS-TCN is also able to recognize phase transitions better than TeCNO in some instances (e.g. P3, P4, P9). Additionally, both the methods outperform ResNet and ResNetLSTM models.

Fig. \ref{fig:step_predictions} visualizes the complete video set of one best and one worst performance of MTMS-TCN for step recognition. Since there are 44 steps, visualizing all of them is quite challenging and clutters the plot. To effectively show the results, we look at one videos instead of 3 in each best and worst category. Furthermore, for better visualization we use a 20 categorical colormap and all 44 steps are mapped onto this colormap. The results clearly show that MTMS-TCN is able to better capture smaller steps and step transitions in comparison to TeCNO and ResNetLSTM. 

\section{Conclusion}

In this paper, we introduce a new multi-level surgical activity annotations for the LRYGB procedures, namely phases and steps. We proposed MTMS-TCN, a multi-task multi-stage temporal convolutional network that was successfully deployed for joint online phase and step recognition. The model is evaluated on a new dataset and compared to state-of-the-art methods in both single-task and multi-task setups and demonstrates the benefits of modeling jointly the phases and steps for surgical workflow recognition. 
\\ \\
{
{\bfseries Acknowledgements} This work has received funding from the European Union's Horizon 2020 research and innovation programme under the Marie Sklodowska-Curie grant agreement No 813782 - project ATLAS. This work was also supported by French state funds managed within the Investissements d'Avenir program by BPI France (project CONDOR) and by the ANR (ANR-16-CE33-0009, ANR-10-IAHU-02). The authors would also like to thank the IHU and IRCAD research teams for their help with the data annotation during the CONDOR project. 
%

}
\bibliographystyle{spmpsci}      
\bibliography{reference}
\end{document}